\documentclass{article}

\usepackage{arxiv}
\usepackage[colorlinks=true,linkcolor=blue,citecolor=red]{hyperref}
\usepackage{booktabs}       
\usepackage{amsfonts}       
\usepackage{nicefrac}       
\usepackage{microtype}      
\usepackage{graphicx}
\usepackage{float}
\usepackage{dsfont}
\usepackage{pgfgantt}
\usepackage{xcolor}
\usepackage{amssymb}
\usepackage[utf8]{inputenc}
\usepackage{amsmath}
\usepackage{xparse}
\usepackage{tikz}
\renewcommand{\vec}[1]{\mathbf{#1}}
\NewDocumentCommand{\tens}{e{_^}}{%
  \mathbin{\mathop{\otimes}\displaylimits
    \IfValueT{#1}{_{#1}}
    \IfValueT{#2}{^{#2}}
  }%
}

\title{Trying Bilinear Pooling in Video-QA}
\author{
  Thomas Winterbottom*\\
  Department of Computer Science\\
  Durham University\\
  United Kingdom \\
  \texttt{thomas.i.winterbottom@durham.ac.uk} \\
   \And
 Sarah Xiao \\
  Business School\\
  Durham University\\
  United Kingdom \\
  \texttt{hong.xiao@durham.ac.uk} \\
   \And
 Alistair McLean \\
  Carbon (AI)\\
  Middlesbrough\\
  United Kingdom \\
  \texttt{alistair@carbonrmp.com} \\
   \And
 Noura Al Moubayed \\
  Department of Computer Science\\
  Durham University\\
  United Kingdom \\
  \texttt{noura.al-moubayed@durham.ac.uk} \\
}

\begin{document}
\maketitle

\begin{abstract}
Bilinear pooling (BLP) refers to a family of operations recently developed for fusing features from different modalities predominantly developed for VQA models. A bilinear (outer-product) expansion is thought to encourage models to learn interactions between two feature spaces and has experimentally outperformed `simpler' vector operations (concatenation and element-wise-addition/multiplication) on VQA benchmarks. Successive BLP techniques have yielded higher performance with lower computational expense and are often implemented alongside attention mechanisms. However, despite significant progress in VQA, BLP methods have not been widely applied to more recently explored video question answering (video-QA) tasks. In this paper, we begin to bridge this research gap by applying BLP techniques to various video-QA benchmarks, namely: TVQA, TGIF-QA, Ego-VQA and MSVD-QA. We share our results on the TVQA baseline model, and the recently proposed heterogeneous-memory-enchanced multimodal attention (HME) model. Our experiments include both simply replacing feature concatenation in the existing models with BLP, and a modified version of the TVQA baseline to accommodate BLP we name the `dual-stream' model. We find that our relatively simple integration of BLP does not increase, and mostly harms, performance on these video-QA benchmarks. Using recently proposed theoretical multimodal fusion taxonomies, we offer insight into why BLP-driven performance gain for video-QA benchmarks may be more difficult to achieve than in earlier VQA models. We suggest a few additional `best-practices' to consider when applying BLP to video-QA. We stress that video-QA models should carefully consider where the complex representational potential from BLP is actually needed to avoid computational expense on `redundant' fusion.
\end{abstract}

\keywords{Video Question Answering \and Bilinear Pooling \and Deep-CCA \and Multimodality \and TVQA \and Ego-VQA \and MSVD-QA \and TGif-QA.}

\section{Introduction}

Advances in deep neural networks over the past decade has allowed models to begin solving new tasks requiring information from more than modality. The overlap of language and vision has a key area of research, in particular visual question answering (VQA) \cite{Agrawal2015VQAVQ, Gupta2017SurveyOV} i.e. answer a question about an image (surveyed here \cite{Wu2017VisualQA, Srivastava2019VisualQA}). However, models solving human tasks will need to look beyond the semantic information in text and visuals and exhibit nuanced behaviour: learn complex referential relations with many cases and exceptions (e.g. `green' may mean a plant is healthy, or a person is ill, inexperienced, envious or recycling), or use one modality to look beyond learned statistical biases in the other \cite{Agrawal2018DontJA, Cadne2019RUBiRU, Chen2020CounterfactualSS, goyal2017making, Ramakrishnan2018OvercomingLP} (`what colour is the banana?', we understand that usually they are yellow, but the one in a particular image may be pink). Representing such interactions poses a serious challenge in deep learning as it is not immediately obvious how to best fuse features tensors in a network. Early works use vector concatenation to project different features into a new joint feature space \cite{Zhou2015SimpleBF, Lu2016HierarchicalQC}. Inspired by an earlier vision-only two-factor framework \cite{Tenenbaum:2000:SSC:1121517.1121518}, VQA models improved significantly by adopting a `pooled' bilinear representation (BLP) of vision and text features \cite{DBLP:journals/corr/GaoBZD15, DBLP:journals/corr/FukuiPYRDR16, DBLP:journals/corr/KongF16, DBLP:journals/corr/KimOKHZ16, DBLP:journals/corr/abs-1708-01471, ben2019block, Yu2018BeyondBG}. Research into BLP for VQA has focused on: creating more effective and less expensive bilinear representations \cite{DBLP:journals/corr/KongF16, DBLP:journals/corr/KimOKHZ16, DBLP:journals/corr/abs-1708-01471, ben2019block}, integrating BLP with attention mechanisms \cite{Kim2018BilinearAN, Yu2018BeyondBG, DBLP:journals/corr/abs-1708-01471}, and recently analysing and categorising BLP (as a ``joint" representation) alongside other fusion techniques and models \cite{Baltruaitis2019MultimodalML, zhang2019multimodal, 8715409}. Despite successes on VQA, BLP techniques have yet to be widely applied to video-QA baselines \cite{MovieQA, pororoqa, jang-CVPR-2017, Fan_2019_ICCV, lei2018tvqa, Xu2017VideoQA}. In this paper, we begin bridging this research gap by applying BLP techniques to the TVQA \cite{lei2018tvqa}, TGIF-QA \cite{jang-CVPR-2017}, Ego-VQA \cite{Fan_2019_ICCV} and MSVD-QA \cite{Xu2017VideoQA} datasets. Our contributions include: \textbf{I)} Replacing concatenation with BLP techniques on the TVQA baseline model \cite{lei2018tvqa}, \textbf{II)} Experimenting with a modified version of the TVQA baseline to accommodate BLP we name the `dual-stream' model, \textbf{III)} Experiments replacing concatenation with BLP on the recently proposed heterogenous-memory-enhanced multimodal attention (HME) model \cite{Fan2019HeterogeneousME}, \textbf{IV)} Using the TVQA dataset in our experiments on HME, \textbf{V)} An insight into the poor performance with respect to recent multimodal fusion taxonomies \cite{Baltruaitis2019MultimodalML, zhang2019multimodal, 8715409}, \textbf{VI)} Experiments on the TVQA baseline model augmented with deep canonical correlation analysis (DCCA) \cite{Andrew2013DeepCC} to contrast a `co-ordinated' representation with BLP's `joint' representation (as defined in \cite{8715409}), \textbf{VII)} A few additional best practices to consider when transitioning BLP from image-QA to video-QA, \textbf{VIII)} An overview of how bilinear representations compare and contrast with current psychological vision and perception models \cite{Goodale1992SeparateVP, milner2006visual, milner2008two, goodale2014and, Milner2017HowDT}. Infamously, inconclusive or negative results of experiments often remain unpublished, a practice that has been widely criticised \cite{Mlinari2017DealingWT, Faschinger2019DoNH, Sandercock2012NegativeRW}. We wish to report our experimental results to expand the public experimental scope of multimodal fusion and encourage critical future research that may be able to build on, contrast and criticise our findings and speculations.

\section{Related Works}
In this section, we outline: \textbf{I)} The development of bilinear methods in multimodal tasks, \textbf{II)} Recent video-QA datasets and their benchmark models, \textbf{III)} Video models that use BLP.

\subsection{Vector Concatenation}
Early works use Vector concatenation to project different features into a new joint feature space. \cite{Zhou2015SimpleBF} use vector concatenation on the CNN image and text features in their simple baseline VQA model. Similarly, \cite{Lu2016HierarchicalQC} concatenate image attention and textual features. Vector concatenation is a projection of both input vector into a new `joint' dimensional space.

\subsection{Bilinear Models}
Strictly speaking, a function is \textit{bilinear} if it is linear in both input domains, though a bilinear representation here refers to the multiplicative tensor-product expansion of two vectors. Working from the observations that ``perceptual systems routinely separate `content' from `style'", \cite{Tenenbaum:2000:SSC:1121517.1121518} proposed a bilinear framework on these two different aspects of purely visual inputs. They find that the multiplicative bilinear model provide ``sufficiently expressive representations of factor interactions". The bilinear model in \cite{Lin2015BilinearCF} is a `two-stream' architecture where distinct subnetworks model temporal and spatial aspects. The bilinear interactions are between the outputs of two CNN streams, resulting in a bilinear vector that is essentially an outer product directly on convolution maps (features are aggregated somewhat with sum-pooling). This makes intuitive sense as convolution maps will learn various patterns and learnable parameters representing the outer product between these maps should learn visualisable and distinct interactions. Interestingly, both \cite{Tenenbaum:2000:SSC:1121517.1121518, Lin2015BilinearCF} are reminiscent of two-stream hypothesises of visual processing in the human brain \cite{Goodale1992SeparateVP, milner2006visual, milner2008two, goodale2014and, Milner2017HowDT} (discussed in detail later). Though these models focus on only visual content, their generalisable two-factor frameworks would later be inspiration to multimodal representations.

\subsection{Compact Bilinear Pooling}
Gao et al. \cite{DBLP:journals/corr/GaoBZD15} introduce `Compact Bilinear Pooling', a technique combining the count sketch function \cite{10.1007/3-540-45465-9_59} and convolution theorem \cite{7015692} in order to `pool' the outer product into a smaller bilinear representation. \cite{DBLP:journals/corr/FukuiPYRDR16} use compact BLP in their VQA model to learn interactions between text and images i.e. multimodal compact bilinear pooling (MCB). We note that for MCB, the learned outer product is no longer on convolution maps (from previously discussed bilinear models), but rather on the indexes 2048-dimensional image and textual tensors. Intuitively, a given index of an image or textual tensor has less distinct meaning from any other index compared to convolution maps which are theoretically learning interactions directly between patterns. As far as we are aware no research has been done discussing potential ramifications of this, and later usages of bilinear pooling methods continue this trend. Though MCB is significantly more efficient than bilinear full bilinear expansions, they still require relatively large latent dimension to perform well on VQA (\textit{d}=16000). 

\subsection{Multimodal Low-Rank Bilinear Pooling}
To further reduce the number of needed parameters, \cite{DBLP:journals/corr/KimOKHZ16} introduce multimodal low-rank bilinear pooling (MLB), which approximates the outer-product weight representation $W$ by decomposing it into two rank-reduced projection matrices:
\begin{center}
 $\vec{z} = MLB(\vec{x},\vec{y}) = (X^T\vec{x})\odot(Y^T\vec{y})$\\
 $\vec{z} = \vec{x}^TW\vec{y} = \vec{x}^TXY^T\vec{y} = \mathds{1}^T(X^T\vec{x} \odot Y^T\vec{y})$
\end{center}
where $X \in \mathbb{R}^{m \times o}$, $Y \in \mathbb{R}^{n \times o}$, $o<min(m,n)$ is the output vector dimension, $\odot$ is element-wise multiplication of vectors or the Hadamard product and $\mathds{1}$ is the vector of all ones. MLB performs better than MCB in the study from \cite{Osman2018DualRA}, but it is sensitive to hyper-parameters and converges slowly. Furthermore \cite{DBLP:journals/corr/KimOKHZ16} suggest using \textit{Tanh} activation on the output of $\vec{z}$ to further increase model capacity. We note that, strictly speaking, adding the the nonlinearity means the representation is no longer bilinear as it is not linear with respect to either of its input domains. 

\subsection{Multimodal Factorised Low Rank Bilinear Pooling} 
\cite{yu2017multi} propose multimodal factorised bilinear pooling (MFB) as an extension of MLB. Consider the bilinear projection matrix $\vec{W} \in \mathbb{R}^{m \times n}$ as before. To learn output $\vec{z} \in \mathbb{R}^{o}$ we need to learn $\vec{W} = [\vec{W}_0,...,\vec{W}_{o-1}]$. So we generalise output $\vec{z}$:
\begin{equation} \label{eq1}
  z_i = \vec{x}^T\vec{X}_i\vec{Y}_i^T\vec{y} 
      = \sum_{d=0}^{k-1}\vec{x}^Ta_db_d^T\vec{y}
      = \mathds{1}^T(\vec{X}_i^T\vec{x} \odot \vec{Y}_i^T\vec{y})
\end{equation} 
Note that MLB is the case of MFB where \textit{k}=1. MFB can be thought of as a two-part process: features are `expanded' to higher-dimensional space by $\vec{W}_{\sigma}$ matrices, then `squeezed' into a ``compact ouput''. The authors argue that this gives ``more powerful'' representational capacity in the same dimensional space than MLB.

\subsection{Multimodal Tucker Fusion}
\cite{Benyounes2017MUTANMT} extend the rank-reduction concept from MLB and MFB to factorise the entire bilinear tensor using tucker decomposition in their multimodal tucker fusion (MUTAN) model \cite{tucker1966some}.

\subsubsection{Rank and mode-n product}
We note that conventionally, the mode-n fibres count from 1 instead of 0. We will follow this convention for the tensor product portion of our paper to avoid confusion. If $\vec{W} \in \mathbb{R}^{I_1 \times,...,\times I_{N}}$ and $\vec{V} \in \mathbb{R}^{J_n \times I_{n}}$ for some
$n \in \{1,...,N\}$ then
\begin{center}
  rank($\vec{W} \tens_{n} \vec{V}$) $\leq$ rank($\vec{W}$)
\end{center}
where $\tens_{n}$ is the mode-n tensor product:
\begin{center}
  ($\vec{W} \tens_{n} \vec{V}$)($i_1,...,i_{n-1},j_n,i_{n+1},...,i_{N}$):=$\sum_{i_n=1}^{I_n}\vec{W}$
  ($i_1,...,i_{n-1},i_n,i_{n+1},...,i_{N}$)$\vec{V}$($j_n,i_n$)
\end{center}
In essence, the mode-n fibres (also known as mode-n vectors) of $\vec{W} \tens_{n} \vec{V}$ are the mode-n fibres of $\vec{W}$ multiplied by $\vec{V}$ (Proof on page 11 here \cite{ddissi}). Each mode-n tensor product introduces an upper bound to the rank of the tensor.
\begin{figure}[ht]
  \centering
   \includegraphics[width=0.6\columnwidth]{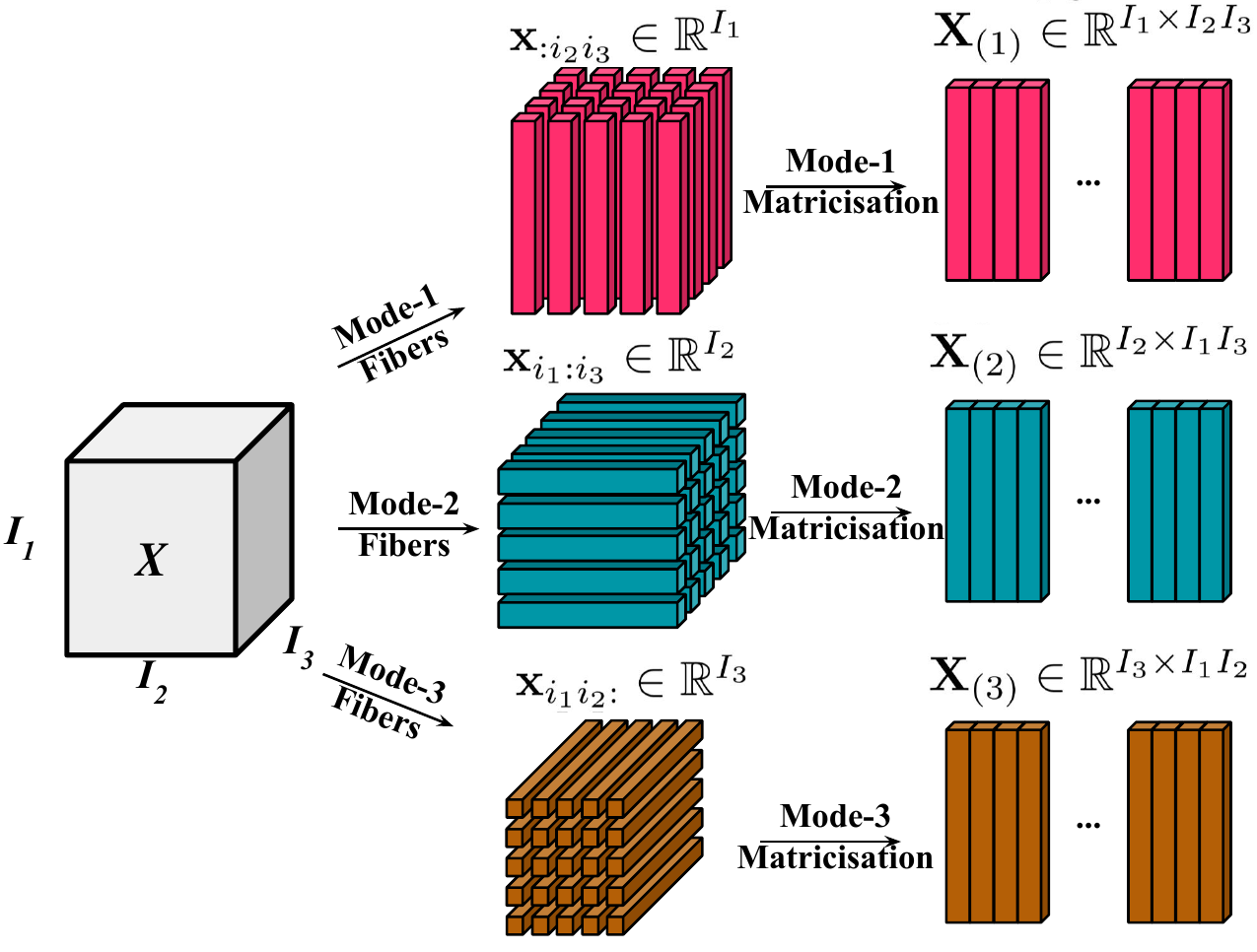}
  \caption{Visualisation of mode-n fibres and matricisation \cite{imgsdsa}}
\end{figure}

\subsubsection{Tucker Decomposition Model}
The tucker decomposition of a real $3^{rd}$ order tensor $\vec{T} \in \mathbb{R}^{d_{1} \times d_{2} \times d_{3}}$ is:
\begin{center}
  $\vec{T} = \tau \tens_{1} \vec{W}_{1} \tens_{2} \vec{W}_{2} \tens_{3} \vec{W}_{3}$
\end{center}
where $\tau \in \mathbb{R}^{d_{1} \times d_{2} \times d_{3}}$ \textit{(core tensor)}, and $\vec{W}_{1}, \vec{W}_{2}, \vec{W}_{3} \in \mathbb{R}^{d_{1} \times d_{1}}$, $\mathbb{R}^{d_{2} \times d_{2}}$, $\mathbb{R}^{d_{3} \times d_{3}}$ \textit{(factor matrices)} respectively. The MUTAN model uses a reduced rank on the core tensor to constrain representational capacity, and the factor matrices to encode full bilinear projections of the textual and visual features, and finally output an answer prediction, i.e:
\begin{center}
    $\vec{y} = ((\tau \tens_{1} (\vec{q}^{T}\vec{W}_{q})) \tens_{2} (\vec{v}^{T}\vec{W}_{v})) \tens_{3} \vec{W}_{o}$
\end{center}
Where $\vec{y} \in \mathbb{R}^{|A|}$ is the answer prediction vector and $\vec{q}, \vec{v}$ are the textual and visual features respectively. A slice-wise attention mechanism is used in the MUTAN model to focus on the most discriminative interactions.
\subsection{Multimodal Factorised Higher Order Bilinear Pooling}
\cite{Yu2018BeyondBG} propose multimodal factorised higher-order bilinear pooling (MFH), extending second-order bilinear pooling to `generalised high-order pooling' by stacking multiple MFB units, i.e:
\begin{center}
    $\vec{z}_{exp}^i = MFB^i_{exp}(\vec{I},\vec{Q}) = \vec{z}_{exp}^{i-1} \odot Dropout(\vec{U}^T\vec{I} \odot \vec{V}^T\vec{Q})$\\
    $\vec{z} = SumPool(\vec{z}_{exp})$
\end{center}
for $i \in \{1,...,p\}$ where $\vec{I}$, $\vec{Q}$ are visual and text features respectively. Similar to how MFB extends MLB, MFH is MFB where $p=1$.

\subsection{Bilinear Superdiagonal Fusion}
\cite{ben2019block} proposed another method of rank restricted bilinear pooling: Bilinear Superdiagonal Fusion (BLOCK). 

\subsubsection{Block Term Decomposition}
Introduced in a 3-part paper \cite{de2008decompositions,de2008decompositions2,de2008decompositions3}, block term decomposition reformulates a bilinear matrix representation as sum of rank restricted matrix products (contrasting low rank pooling which is represented by only a single rank restricted matrix product). By choosing the number of decompositions in the approximated sum and their rank, block-term decompositions offer greater control over the approximated bilinear model. Block term decompositions are easily extended to higher-order tensor decompositions, allowing multilinear rank restriction for multilinear models in future research. A \textit{block term decomposition} of a tensor $\vec{W} \in \mathbb{R}^{I_1 \times,...,\times I_{N}}$ is a decomposition of the form: 
\begin{center}
$\vec{W}=\sum_{r=1}^{R}\vec{S}_r\tens_{1}\vec{U}_r^1\tens_{2}\vec{U}_r^2\tens_{3},...,\tens_{n}\vec{U}_r^{n}$
\end{center}
where $R \in \mathbb{N}^*$ and for each $r \in \{1,...,R\}, \vec{S}_r \in \mathbb{R}^{R_1 \times ,..., \times R_{n}}$ where each $\vec{S}_r$ are `core tensors' with dimensions $R_n \leq I_n$ for $n \in \{1,...,N\}$ that are used to restrict the rank of the tensor $\vec{W}$. $\vec{U}_r^n \in$ St($R_n,I_n$) are the `factor matrices' that intuitively expand the \textit{nth} dimension of $\vec{S}$ back up to the original \textit{nth} dimension of $\vec{W}$. St($a,b$) here refers to the Stiefel manifold, i.e. St($a,b$):$\{\vec{Y} \in \mathbb{R}^{a \times b}:\vec{Y}^T\vec{Y}=\vec{I}_p\}$.\\

\begin{figure}[ht]
  \centering
   \includegraphics[width=0.7\columnwidth]{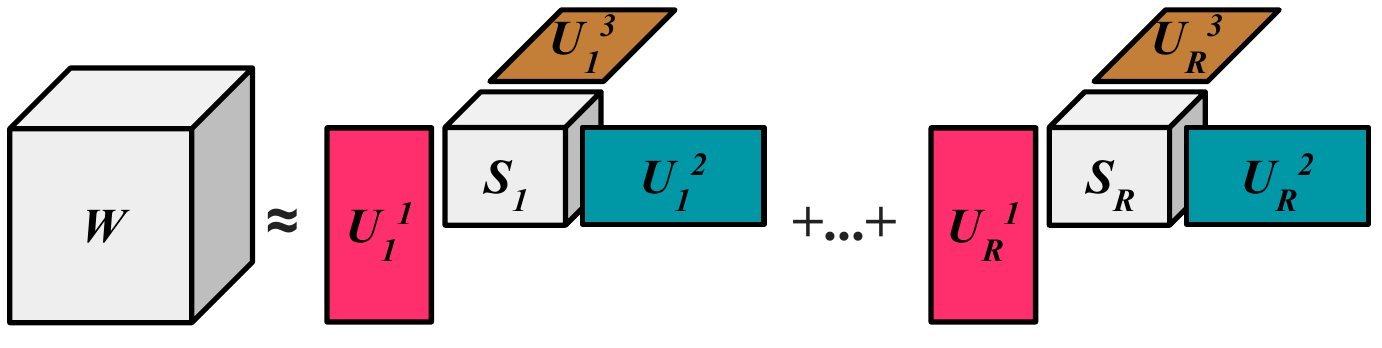}
  \caption{Block Term Decomposition (n=3)}
\end{figure}

\subsubsection{Bilinear Superdiagonal Model}
The BLOCK model uses block term decompositions to learn multimodal interactions. The authors argue that since BLOCK enables ``very rich (full bilinear) interactions between groups of features, while the block structure limits the complexity of the whole model'', that it is able to represent very fine grained interactions between modalities while maintaining powerful mono-modal representations. The bilinear model with inputs $\vec{x} \in \mathbb{R}^m, \vec{y} \in \mathbb{R}^n$ is projected into \textit{o} dimensional space with tensor products:
\begin{center}
  $\vec{z} = \vec{W} \tens_{1} \vec{x} \tens_{2} \vec{y}$
\end{center}
where $\vec{z} \in \mathbb{R}^o$. The superdiagonal BLOCK model uses a 3 dimensional block term decomposition. The decomposition of $\vec{W}$ in rank ($R_1,R_2,R_3$) is defined as:
\begin{center}
  $\vec{W}=\sum_{r=1}^{R}\vec{S}_r\tens_{1}\vec{U}_r^1\tens_{2}\vec{U}_r^2\tens_{3}\vec{U}_r^3$
\end{center}
This can be written as
\begin{center}
  $\vec{W}=\vec{S}^{bd}\tens_{1}\vec{U}^1\tens_{2}\vec{U}^2\tens_{3}\vec{U}^{3}$
\end{center}
where $\vec{U}^1 = $[$\vec{U}_1^1,...,\vec{U}_{R}^1$], similarly with $\vec{U}^2$ and $\vec{U}^3$, and now $\vec{S}^{bd} \in \mathbb{R}^{RR^1 \times RR^2 \times RR^3}$. So $\vec{z}$ can now be expressed with respect to $\vec{x}$ and $\vec{y}$. Let $\hat{\vec{x}}=\vec{U}^1\vec{x} \in \mathbb{R}^{RR^1}$ and $\hat{\vec{y}}=\vec{U}^2\vec{y} \in \mathbb{R}^{RR^2}$. These two projections are merged by the block-superdiagonal tensor $\vec{S}^{bd}$. Each block in $\vec{S}^{bd}$  merges together blocks of size $R^1$ from $\hat{\vec{x}}$ and of size $R^2$ from $\hat{\vec{y}}$ to produce a vector of size $R^3$: 
\begin{center}
  $\vec{z}_r = \vec{S}_r \tens_{x} \hat{\vec{x}}_{rR^1:\left(r+1\right)R^1} \tens_{y} \hat{\vec{y}}_{rR^2:\left(r+1\right)R^2}$
\end{center}
where $\hat{\vec{x}}_{i:j}$ is the vector of dimension $j-i$ containing the corresponding values of $\hat{\vec{x}}$. Finally all vectors $\vec{z}_r$ are concatenated producing $\hat{\vec{z}} \in \mathbb{R}^{RR^3}$. The final prediction vector is $\vec{z}=\vec{U}^3\hat{\vec{z}} \in \mathbb{R}^o$. 

\subsection{BLP in Video Datasets}
Though we aim to address a research gap specifically considering the role of BLP in video benchmarks, several recent video models have incorporated and contrasted BLP techniques to their own model designs. \cite{Kim2019ProgressiveAM} find various BLP fusions perform worse than their `dynamic modality fusion' mechanism on TVQA \cite{lei2018tvqa} and MovieQA \cite{MovieQA}. \cite{Li2019LearnableAN} consider MCB fusion on their ablation studies in TGIF-QA \cite{jang-CVPR-2017}. \cite{Chou2020VisualQA} use MLB as part of their baseline model proposed alongside their `VQA 360$^{\circ}$' dataset. \cite{Gao2019StructuredTA} contrast their proposed two-stream attention mechanism to an MCB model for TGIF-QA, demonstrating a substantial performance increase over previous approaches. The Focal Visual-Text Attention network (FVTA) \cite{Liang2019FocalVA} is a hierarchical model that aims to dynamically select from the appropriate point across both time and modalities that outperforms an MCB approach on Movie-QA. The success of these newer models indicate that there is much to gain from a more nuanced approach to multimodal fusion in video-QA. However, in this paper we aim to consider a more targeted and contrastive analysis of BLP as a multimodal fusion device.

\section{Datasets}

\subsection{MSVD-QA}
\cite{Xu2017VideoQA} argue that simply extending image-QA methods is ``insufficient and suboptimal'' to conduce quality video-QA, and that instead the focus should be the temporal structure of videos. Using an NLP method to automatically generate QA pairs automatically from descriptions \cite{Heilman2009QuestionGV}, \cite{Xu2017VideoQA} create the MSVD-QA dataset based on the Microsoft research video description corpus \cite{Chen2011CollectingHP}. The dataset is made from 1970 video clips, with over 50k QA pairs in `5w' style i.e. ("what", "who", "how", "when", "where").

\subsection{TGIF-QA}
\cite{jang-CVPR-2017} speculate that the relatively limited progress in video-QA compared to image-QA is ``due in part to the lack of large-scale datasets with well defined tasks''. As such, they introduced the TGIF-QA dataset to `complement rather than compete' with existing VQA literature to serves as a bridge between video-QA and video understanding. To this end, they propose 3 subsets with specific video-QA tasks that specifically take advantage of the temporal format of videos:

\noindent \textbf{Count:} Counting the number of times a specific action is repeated \cite{Levy2015LiveRC} e.g. ``How many times does the girl jump?''. Models output the predicted number of times the specified actions happened. (Over 30k QA pairs).

\noindent \textbf{Action:} Identify the action that is repeated a number of times in the video clip. There are over 22k multiple choice questions e.g. ``What does the girl do 5 times?''.

\noindent \textbf{Trans:} Identifying details about a state transition (\cite{Isola2015DiscoveringSA}). There are over 58k multiple choice questions e.g. ``What does the man do after the goal post?".

\noindent \textbf{Frame-QA:} An image-QA split using automatically generated QA pairs from frames and captions in the TGIF dataset \cite{Li2016TGIFAN} (over 53k multiple choice questions).

\subsection{TVQA}
The TVQA dataset \cite{lei2018tvqa} is designed to address the shortcomings of previous datasets. It has significantly longer clip lengths than other datasets and is based on TV shows instead of cartoons, giving it realistic video content with simple coherent narratives. It contains over 150k QA pairs. Each question is labelled with timestamps for the relevant video frames and subtitles. The questions were gathered using AMT workers. Most notably, the questions were specifically designed to encourage multimodal reasoning by asking the workers to design two-part compositional questions. The first part asks a question about a `moment' and the second part localises the relevant moment in the video clip i.e. [What/How/Where/Why/Who/...] --- [when/before/after] ---, e.g. \textit{[What] was House saying [before] he leaned over the bed?}. The authors argue this facilitates questions that require both visual and language information since ``people often naturally use visual signals to ground questions in time''. The authors identify certain biases in the dataset. They find that the average length of correct answers are longer than incorrect answers. They analyse the performance of their proposed baseline model with different combinations of visual and textual features on different question types they have identified. Though recent analysis has highlighted bias towards subtitles in TVQA's questions \cite{mbintvqa}, it remains an important large scale video-QA benchmark.

\subsection{Ego-VQA}
Most video-QA datasets focus on video-clips from the $3^{rd}$ person. \cite{Fan_2019_ICCV} argue that $1^{st}$ person video-QA has more natural use cases that real-world agents would need. As such, \cite{Fan_2019_ICCV} propose the egocentric video-QA dataset (Ego-VQA) with 609 QA pairs on 16 first-person video clips. Though the dataset is relatively small, it has a diverse set of question types (e.g. $1^{st}$ \& $3^{rd}$ person `action' and `who' questions, `count', `colour' etc..), and generates hard and confusing incorrect answers by sampling from correct answers of the same question type. Models on Ego-VQA tend to overfit due to its small size, to remedy this, \cite{Fan_2019_ICCV} pretrain the baseline models on the larger YouTube2Text-QA \cite{DBLP:journals/corr/YeZLCXZ17}. YouTube2Text-QA is a multiple choice dataset created from MSVD videos \cite{Chen2011CollectingHP} and questions created from YouTube2Text video description corpus \cite{Guadarrama2013YouTube2TextRA}. YouTube2Text-QA has over 99k questions in `what', `who' and `other' style. We believe Ego-VQA represents an interesting new approach to video datasets. We believe that researchers should not be discouraged by the often steep price tag of collecting and annotating large scale datasets, and should look to create smaller innovative datasets since larger more conventional video-QA datasets now exist for pretraining.

\section{Models}
We build our models from official TVQA \footnote{https://github.com/jayleicn/TVQA} and HME-VideoQA \footnote{https://github.com/fanchenyou/HME-VideoQA} implementations.
\subsection{TVQA Model}
\begin{figure*}
    \centering
    \includegraphics[width=1.0\textwidth]{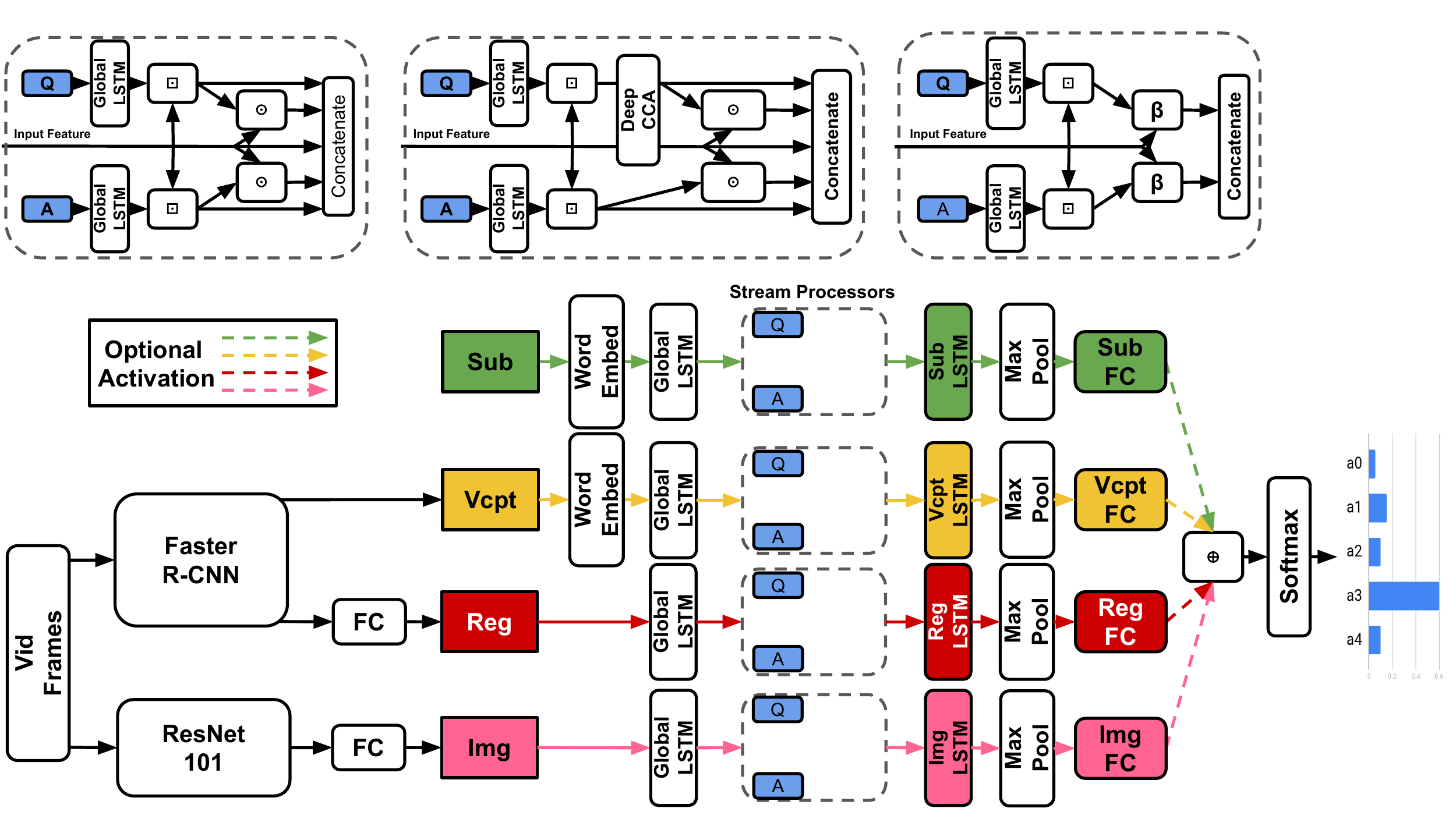}
    \caption{TVQA Model. $\odot$/$\oplus$ = Element-wise multiplication/addition, $\boxdot$ = context matching \cite{DBLP:journals/corr/SeoKFH16, DBLP:journals/corr/abs-1804-09541}. Any feature streams may be enabled/disabled.}
    \label{model}
\end{figure*}

\noindent \textbf{Model Definition}: The model takes as inputs, I) A question \textit{q} (13.5 words on average), II) Five potential answers $\{\textit{a}_{i}\}_{i=0}^{4}$ (each between 7-23 words), III) A subtitle \textit{S} and video-clip \textit{V} ($\sim$60-90s at 3FPS), and outputs the predicted answer. As the model can either use the entire video-clip and subtitle or only the parts specified in the timestamp, we refer to the sections of video and subtitle used as segments from now on. Figure \ref{model} demonstrates the textual and visual streams and their associated features in model architecture.

\noindent \textbf{ImageNet Features:} Each frame is processed by a ResNet101  \cite{DBLP:journals/corr/HeZRS15} pretrained on ImageNet \cite{5206848} to produce a 2048-d vector. These vectors are then L2-normalised and stacked in frame order: $V^{img}\in\mathbb{R}^{f\times2048}$ where \textit{f} is the number of frames used in the video segment.

\noindent \textbf{Regional Features:} Each frame is processed by a Faster R-CNN \cite{DBLP:journals/corr/RenHG015} trained on Visual Genome \cite{krishnavisualgenome} in order to detect objects. Each detected object in the frame is given a bounding box, and has an affiliated 2048-d feature extracted. Since there are multiple objects detected per frame (we cap it at 20 per frame), it is difficult to efficiently represent this in time sequences \cite{lei2018tvqa}. The model uses the top-K regions for all detected labels in the segment as in \cite{DBLP:journals/corr/AndersonHBTJGZ17} and \cite{Karpathy2015DeepVA}. Hence the regional features are $V^{reg}\in\mathbb{R}^{n_{reg}\times2048}$ where $n_{reg}$ is the number of regional features used in the segment. 

\noindent \textbf{Visual Concepts:} The classes or labels of the detected regional features are called `Visual Concepts'. \cite{inproceedings223} found that simply using detected labels instead of image features gives comparable performance on image captioning tasks. Importantly they argued that combining CNN features with detected labels outperforms either approach alone. Visual concepts are represented as either GloVe \cite{pennington2014glove} or BERT \cite{devlin2018bert} embeddings $V^{vcpt}\in\mathbb{R}^{n_{vcpt}\times 300}$ or $\mathbb{R}^{n_{vcpt}\times 768}$ respectively, where $n_{vcpt}$ is the number of visual concepts used in the segment.

\noindent \textbf{Text Features:} In the evaluation framework, the model encodes the questions, answers, and subtitles using either GloVe  ($\in\mathbb{R}^{300}$) or BERT embeddings ($\in\mathbb{R}^{768}$). Formally,  $q\in\mathbb{R}^{n_q\times d},  \{\textit{a}_{i}\}_{i=0}^{4}\in\mathbb{R}^{n_{a_i}\times d}, S\in\mathbb{R}^{n_s\times d}$ where $n_q, n_{a_i}, n_s$ is the number of words in $q , a_i, S$ respectively and $d=300, 768$ for GloVe or BERT embeddings respectively.

\noindent \textbf{Context Matching:} Context matching refers to context-query attention layers recently adopt-
ed in machine comprehension \cite{DBLP:journals/corr/SeoKFH16, DBLP:journals/corr/abs-1804-09541}. Given a context-query pair, context matching layers return `context aware queries'.

\noindent \textbf{Model Details:} 
In our evaluation framework, any combination of subtitles or visual features can be used. All features are mapped into word vector space through a tanh non-linear layer. They are then processed by a shared bi-directional LSTM \cite{Hochreiter:1997:LSM:1246443.1246450, graves2005framewise} (`Global LSTM' in Figure \ref{model}) of output dimension 300. Features are context-matched with the question and answers. The original context vector is then concatenated with the context-aware question and answer representations and their combined element-wise product (`Stream Processor' in Figure \ref{model}, e.g. for subtitles \textit{S}, the stream processor outputs [$F^{sub}$;$A^{sub,q}$;$A^{sub,a_{0-4}}$;
$F^{sub}\odot A^{sub,q}$;$F^{sub}\odot A^{sub,a_{0-4}}$]$\in\mathbb{R}^{n_{sub}\times1500}$ where $F^{sub}\in\mathbb{R}^{n_{s}\times300}$. Each concatenated vector is processed by their own unique bi-directional LSTM of output dimension 600, followed by a pair of fully connected layers of output dimensions 500 and 5, both with dropout 0.5 and ReLU activation. The 5-dimensional output represents a vote for each answer. The element-wise sum of each activated feature stream is passed to a softmax producing the predicted answer ID. All features remain separate through the entire network, effectively allowing the model to choose the most useful features.

\subsection{HME-VideoQA}
\begin{figure*}[ht]
    \centering
    \includegraphics[width=1.0\textwidth]{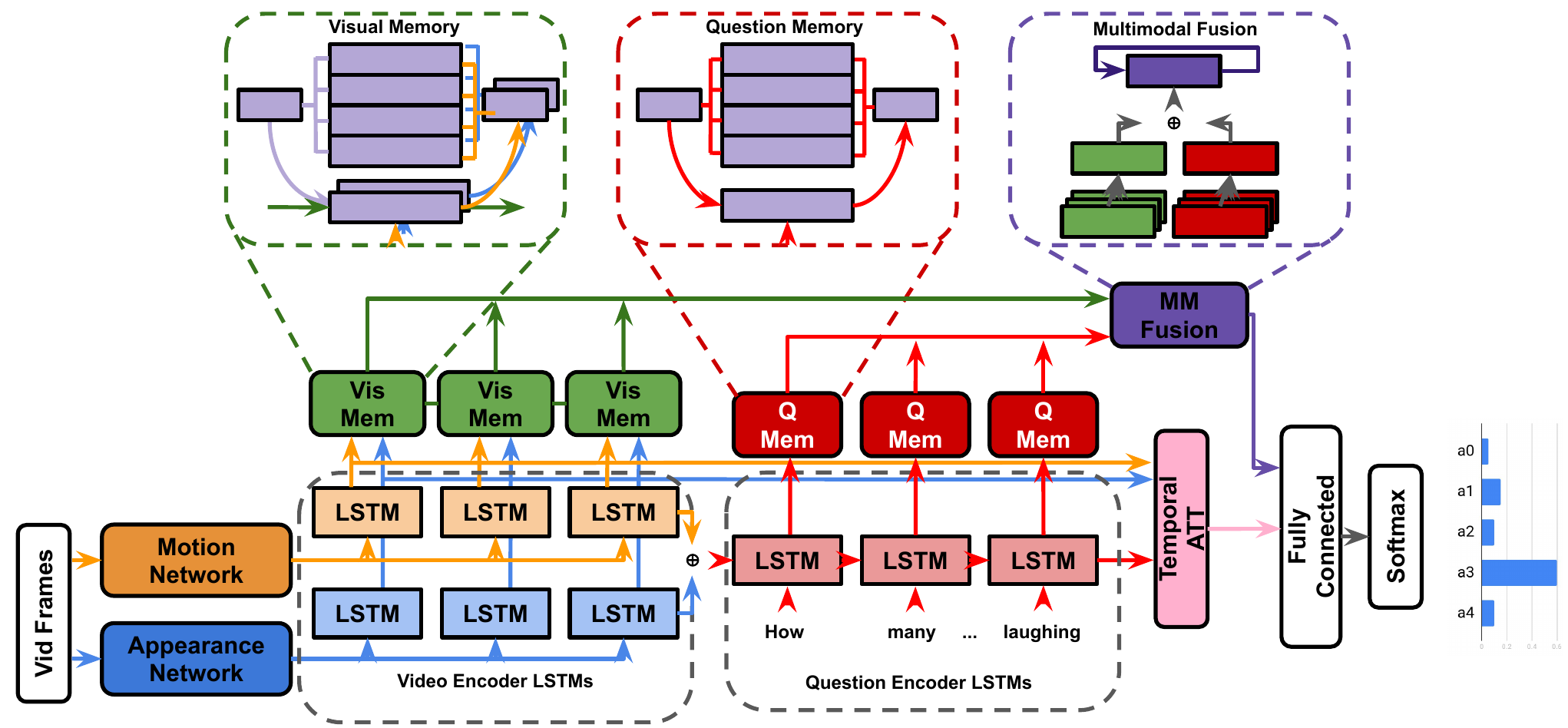}
    \caption{HME Model}
    \label{hme_model}
\end{figure*}

To better handle semantic meaning through long sequential video data, recent models have integrated external `memory' units \cite{Xiong2016DynamicMN, Sukhbaatar2015EndToEndMN} alongside recurrent networks to handle input features \cite{Gao2018MotionAppearanceCN, Zeng2017LeveragingVD}. These external memory units encourage multiple iterations of inference between questions and video features, helping the model revise it's visual understanding as new details from the question are presented. The heterogeneous memory-enhanced video-QA model (HME) \cite{Fan2019HeterogeneousME} proposes several improvements to previous memory based architectures:

\noindent \textbf{Heterogeneous Read/Write Memory:} The memory units in HME use an attention-guided read/write mechanism to read from/update memory units respectively (the number of memory slots used is a hyperparameter). The claim is that since motion and appearance features are heterogeneous, a `straightforward' combination of them cannot effectively describe visual information. The video memory aims to effectively fuses motion (C3D \cite{Tran2014C3DGF}) and appearance (ResNet \cite{DBLP:journals/corr/HeZRS15} and VGG \cite{Simonyan2015VeryDC}) features by integrating them in the joint read/write operations (visual memory in Figure \ref{hme_model}).

\noindent \textbf{Encoder-Aware Question Memory:} Previous memory models used a single feature vector outputted by an LSTM or GRU for their question representation \cite{Gao2018MotionAppearanceCN, Zeng2017LeveragingVD, Xiong2016DynamicMN, DBLP:journals/corr/AndersonHBTJGZ17}. HME use an LSTM question encoder and question memory unit pair that augment eachother dynamically (question memory in Figure \ref{hme_model}).

\noindent \textbf{Multimodal Fusion Unit:} The hidden states of the video and question memory units are processed by a temporal attention mechanism. The joint representation `read' updates the fusion unit's own hidden state. The visual and question representations are ultimately fused by vector concatenation (multimodal fusion in Figure \ref{mm_fusion}). Our experiments will involve replacing this concatenation step with BLP techniques.

\section{Experiments and Results}
In this section we outline our experimental setup and results. We save our insights for the disussion in the next section.
\begin{table}[ht]
  \begin{center}
    \begin{tabular}{r|c|c} 
      \textbf{Dataset} & \textbf{Benchmark} & \textbf{SoTA}\\\hline\hline
      TVQA (Val) & 68.85\% / \cite{lei2018tvqa} & 72.13\% / \cite{mbintvqa}\\
      TVQA (Test) & 68.48\% / \cite{lei2018tvqa} & 70.23\% / \cite{lei2019tvqa}\\\hline
      Ego-VQA (Val 1)  & 37.57\% / \cite{Fan_2019_ICCV} & 45.05*\% / \cite{hmegithub} \\
      Ego-VQA (Test 2)  & 31.02\% / \cite{Fan_2019_ICCV} & 43.35\%* / \cite{hmegithub} \\\hline
      MSVD-QA & 32.00\% / \cite{Xu2017VideoQA} & 36.10\% / \cite{Le2020HierarchicalCR} \\\hline   
      TGIF-Action  & 60.77\% / \cite{jang-CVPR-2017} & 72.38\% / \cite{Gao2019StructuredTA}\\
      TGIF-Count $\dagger$  & 4.28 / \cite{jang-CVPR-2017} & 4.25 / \cite{Gao2019StructuredTA}\\
      TGIF-Trans  & 67.06\% / \cite{jang-CVPR-2017} & 79.03\% / \cite{Gao2019StructuredTA}\\
      TGIF-FrameQA  & 49.27\% / \cite{jang-CVPR-2017} & 56.64\% / \cite{Gao2019StructuredTA}\\
    \end{tabular}
    \caption{Dataset benchmark and SoTA results to the best of our knowledge (excluding this paper). $\dagger$ = Mean L2 loss. * = Replicated results from cited implementation.}
    \label{tab:maintable}
  \end{center}
\end{table}

\subsection{Concatenation to BLP (TVQA) \label{cc2blp}}
\begin{figure}[ht]
    \centering
    \includegraphics[width=0.6\columnwidth]{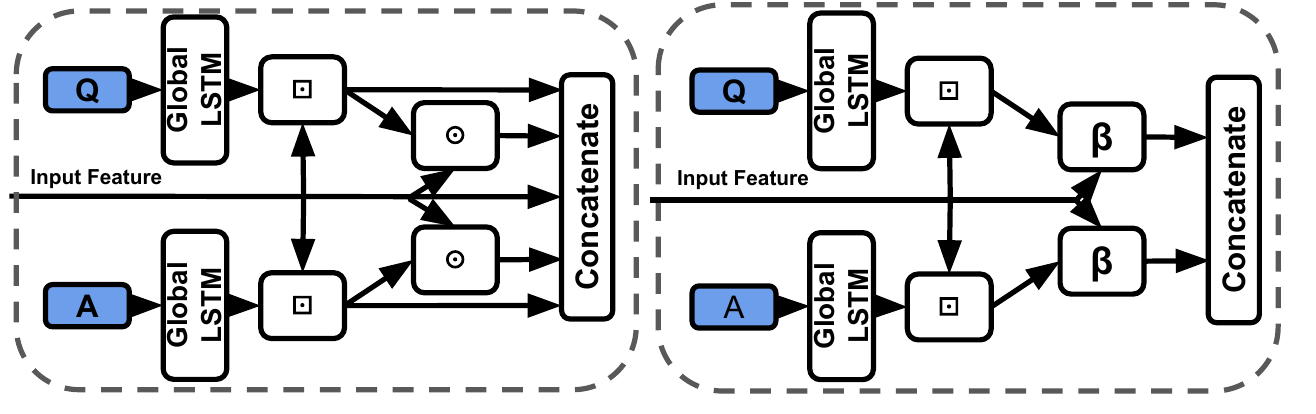}
    \caption{Baseline concatenation stream processor from TVQA model (left) vs Our BLP stream processor (right). $\odot$ = Element-wise multiplication, $\beta$ = BLP, $\boxdot$ = Context Matching.}
    \label{cc2blp_tvqa}
\end{figure}

As previously discussed, BLP techniques have outperformed feature concatenation on a number of VQA benchmarks. The baseline stream processor concatenates the visual feature vector with question and answer representations. Each of the 5 inputs to the final concatenation are 300-d. We replace the visual-question/answer concatenation with BLP (Figure \ref{cc2blp_tvqa}). All inputs to the BLP layer are 300-d, the outputs are 750-d and the hidden size is 1600 (a smaller hidden state that normal, however, the input features are also smaller compared to other uses of BLP). We make as few changes as possible to accommodate BLP, i.e. we use context matching to facilitate BLP fusion by aligning visual and textual features temporally. Our experiments include models with/without subtitles or questions (Table \ref{tab:cc2blp_tvqa}).

\begin{table}[ht]
  \begin{center}
    \begin{tabular}{r|c|c|c} 
      \textbf{Subtitles} & \textbf{Fusion Type} & \textbf{Accuracy} & \textbf{Baseline Offset}\\\hline\hline
      - & Concatenation & 45.94\% & -\\
      GloVE & Concatenation & 69.74\% & -\\
      BERT & Concatenation & 72.20\% & -\\\hline
      - (No Q) & Concatentation & 45.58\% & -0.36\%\\
      GloVE (No Q) & Concatentation & 68.31\% & -1.42\%\\
      BERT (No Q) & Concatentation & 70.43\% & -1.77\%\\\hline
      - & MCB & \textbf{45.65\%} & \textbf{-0.29\%}\\
      GloVE & MCB & \textbf{69.32\%} & \textbf{-0.42\%}\\
      BERT & MCB & \textbf{71.68\%} & \textbf{-0.52\%}\\\hline
      - & MLB & 41.98\% & -3.96\%\\
      GloVE & MLB & 69.30\% & -0.44\%\\
      BERT & MLB & 69.04\% & -3.16\%\\\hline
      - & MFB & 41.82\% & -4.12\%\\
      GloVE & MFB & 68.87\% & -0.87\%\\
      BERT & MFB & 67.29\% & -4.91\%\\\hline
      - & MFH & 44.44\% & -1.5\%\\
      GloVE & MFH & 68.43\% & -1.31\%\\
      BERT & MFH & 67.29\% & -4.91\%\\\hline
      - & Blocktucker & 44.44\% & -1.5\%\\
      GloVE & Blocktucker & 67.95\% & -1.79\%\\
      BERT & Blocktucker & 67.04\% & -5.16\%\\\hline  
      - & BLOCK & 41.09\% & -4.85\%\\
      GloVE & BLOCK & 65.31\% & -4.43\%\\
      BERT & BLOCK & 66.94\% & -5.26\%\\
      
    \end{tabular}
    \caption{Concatenation replaced with BLP in the TVQA model on the TVQA Dataset. All models use visual concepts and ImageNet features.}
    \label{tab:cc2blp_tvqa}
  \end{center}
\end{table}

\subsection{Dual-Stream Model}
\begin{figure}[ht]
    \centering
    \includegraphics[width=\columnwidth]{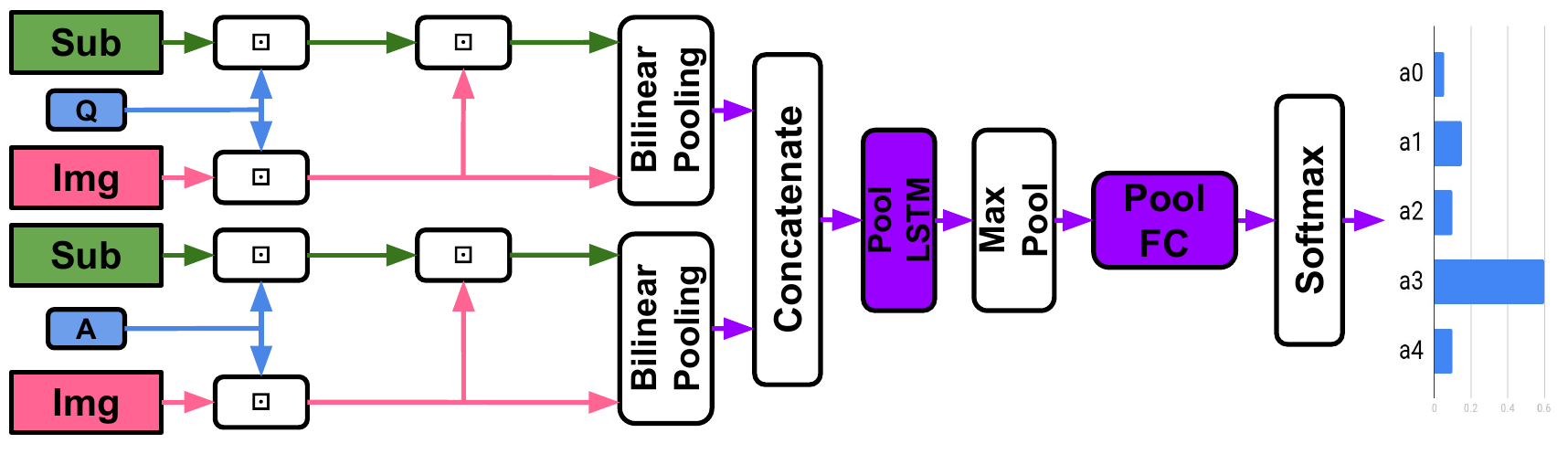}
    \caption{Our Dual-Stream Model. $\boxdot$ = Context Matching.}
    \label{dual_stream_model}
\end{figure}

We create our `dual-stream' (Figure \ref{dual_stream_model}, Table \ref{tab:2stream}) model from the SI TVQA baseline model for 2 main purpose: \textbf{I)} To explore the effects of a joint representation on TVQA, \textbf{II)} To contrast the concatenation-replacement experiment with a model restructured specifically with BLP as a focus. The baseline BLP model keeps subtitles and other visual features completely separate up to the answer voting step. Our aim here is to create a joint representation BLP-based model similar in essence to the baseline TVQA model that fuses subtitle and visual features. As before, we use context matching to temporally align the video and text features. 
\begin{table}[ht]
    \centering
    \begin{tabular}{l|c|c}
      \textbf{Model} & \textbf{Text} & \textbf{Val Acc}\\
      \hline\hline
        TVQA SI & GloVe & 67.78\%\\
        TVQA SI & BERT & 70.56\%\\\hline
        Dual-Stream MCB & GloVe & 63.46\%\\
        Dual-Stream MCB & BERT & 60.63\%\\\hline
        Dual-Stream MFH & GloVe & 62.71\%\\
        Dual-Stream MFH & BERT & 59.34\%\\
    \end{tabular}
    \caption{Dual Stream Results Table.}\label{tab:2stream}
\end{table}

\subsection{Deep CCA in TVQA}
In contrast to joint representations, \cite{Baltruaitis2019MultimodalML} define `co-ordinated representations' as a category of multimodal fusion techniques that learn ``separated but co-ordinated '' representations for each modality (under some constraints). \cite{Peng2018ModalitySpecificCS} claim that since there is often an information imbalance between modalities, learning separate modality representations can be beneficial for preserving `exclusive and useful modality-specific characteristics'. We include one such representation, deep canonical correlation analysis (DCCA) \cite{Andrew2013DeepCC}, in our experiments to contrast with the joint BLP models.

\subsubsection{CCA}
Canonical cross correlation analysis (CCA) \cite{hotelling1936relations} is a method for measuring the correlations between two sets. Let $(\vec{X_0},\vec{X_1}) \in \mathbb{R}^{d_0} \times \mathbb{R}^{d_1}$ be random vectors with covariances $(\sum_{r=00}, \sum_{r=11})$ and cross-covariance $\sum_{r=01}$. CCA finds pairs of linear projections of the two views $(w'_0\vec{X_0}, w_1'\vec{X_1})$ that are maximally correlated:
\begin{center}
    $\rho = (w_0^{*}, w_1^{*}) = \underset{w_{0}, w_{1}}{argmax}$ $corr(w'_0\vec{X_0}, w_1'\vec{X_1})$\\
    $= \underset{w_{0}, w_{1}}{argmax} \frac{w'_{0}\sum_{01}w_{1}}{\sqrt{w'_{0}\sum_{00}w_{0}w'_{1}\sum_{11}w_{1}}}$
\end{center}
where $\rho$ is the correlation co-efficient. As $\rho$ is invariant to the scaling of  $w_{0}$ and $w_{1}$, the projections are constrained to have unit variances, and can be represented as the following maximisation:
\begin{center}
    $\underset{w_{0}, w_{1}}{argmax}$  $w'_{0}\sum_{01}w_{1}$  \textit{s.t}  $w'_{0}\sum_{00}w_{0} = w'_{1}\sum_{11}w_{1} = \vec{1}$
\end{center}
However, CCA can only model linear relationships regardless of the underlying realities in the dataset. Thus, CCA extensions were proposed, including kernel CCA (KCCA) \cite{Akaho2001AKM} and later DCCA.

\subsubsection{DCCA}
DCCA is a parametric method used in multimodal neural networks that can learn non-linear transformations for input modalities. Both modalities $t, v$ are encoded in neural-network transformations $H_t, H_v$ = $f_t(t,\theta_{t})$, $f_v(v,\theta_{v})$ , and then the canonical correlation between both modalities is maximised in a common subspace (i.e. maximise cross-modal correlation between $H_t$, $H_v$).
\begin{center}
    max $corr(H_t, H_v)$ = $\underset{\theta_{t}, \theta_{v}}{argmax}$ $corr(f_t(t,\theta_{t}),f_v(v,\theta_{v}))$
\end{center}
We use DCCA over KCCA to co-ordinate modalities in our experiments as it is generally more stable and efficient, learning more `general' functions.

\subsubsection{DCCA in TVQA}
\begin{figure}[ht]
    \centering
    \includegraphics[width=0.8\columnwidth]{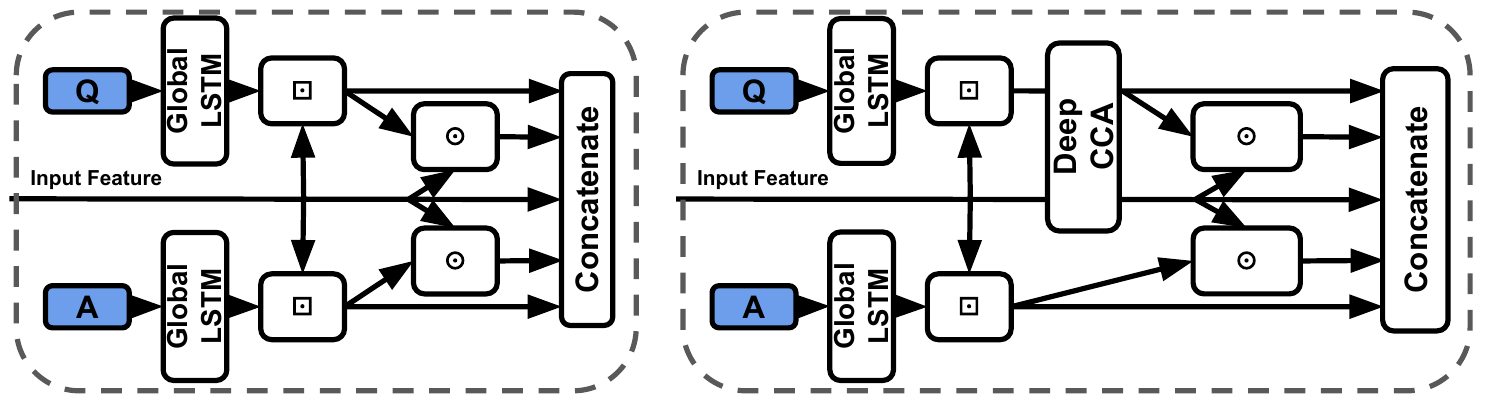}
    \caption{Baseline concatenation stream processor from TVQA model (left) vs Our DCCA stream processor (right). $\odot$ = Element-wise multiplication, $\boxdot$ = Context Matching.}
    \label{fig:dcca}
\end{figure}

We use a 2-layer DCCA module to coordinate question and context (visual or subtitle) features (Figure \ref{fig:dcca}, Table \ref{tab:deepcca}). Output features are the same dimensions as inputs. Though DCCA itself is not directly related to BLP, it has recently been classified as a coordinated representation \cite{8715409}, which is in some respects the opposite of a joint representation.
\begin{table}[ht]
    \centering
    \begin{tabular}{l|c|c|c}
      \textbf{Model} & \textbf{Text} & \textbf{Baseline Acc} & \textbf{DCCA Acc}\\
      \hline\hline
      VI & GloVe & 45.94\% & 45.00\% (-0.94)\\
      VI & BERT & -- & 41.70\% \\\hline
      SVI & GloVe & 69.74\% & 67.91\% (-1.83)\\
      SVI & BERT & 72.20\% & 68.48\% (-3.72)\\
    \end{tabular}
    \caption{DCCA in the TVQA Baseline Model.\label{tab:deepcca}}
\end{table}

\subsection{Concatenation to BLP (HME-VideoQA)}
\begin{figure}[ht]
    \centering
    \includegraphics[width=0.5\columnwidth]{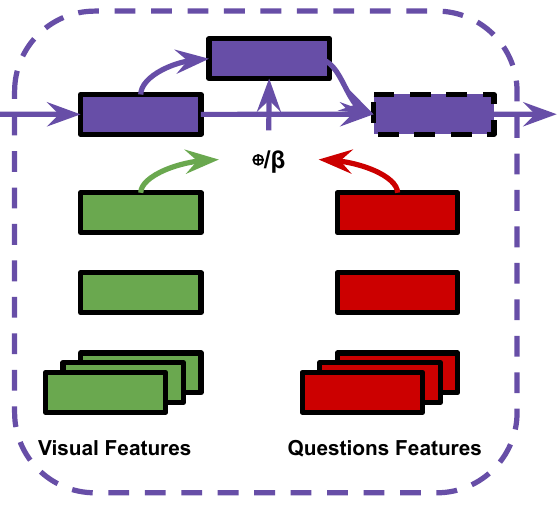}
    \caption{$\oplus$ = Concatenation, $\beta$ = BLP.}
    \label{mm_fusion}
\end{figure}

As in Section \ref{cc2blp}, we replace a concatenation step in the HME model between textual and visual features with BLP (Figure \ref{mm_fusion}, corresponding to the multimodal fusion unit in Figure \ref{hme_model}). The goal here is to explore if BLP can better facilitate multimodal fusion in aggregated memory features (Table \ref{tab:cc2blp_hme}). We replicate the results from \cite{Fan2019HeterogeneousME} with the HME on the MSVD, TGIF and Ego-VQA datasets using the official github repository \cite{hmegithub}. We extract our own C3D features from the frames in the TVQA. To the best of our knowledge, we are the first to apply TVQA to HME.
\begin{table}[ht]
  \begin{center}
    \begin{tabular}{r|c|c|c} 
      \textbf{Dataset} & \textbf{Fusion Type} & \textbf{Val} & \textbf{Test}\\\hline\hline
      TVQA (GloVE) & Concatenation & 41.25\% & N/A\\\hline
      Ego-VQA-0 & Concatenation & 36.99\% & 37.12\%\\
      Ego-VQA-1 & Concatenation & 48.50\% & 43.35\%\\
      Ego-VQA-2 & Concatenation & 45.05\% & 39.04\%\\\hline
      MSVD-QA & Concatenation & 30.94\% & 33.42\%\\\hline
      TGIF-Action & Concatenation & 70.69\% & 73.87\%\\
      TGIF-Count & Concatenation & 3.95$\dagger$ & 3.92$\dagger$\\
      TGIF-Trans & Concatenation & 76.33\% & 78.94\%\\
      TGIF-FrameQA & Concatenation & 52.48\% & 51.41\%\\\hline\hline
      TVQA (GloVE) & MCB & 41.09\% (-0.16) & N/A\%\\\hline
      Ego-VQA-0 & MCB & No Convergence & No Convergence \\
      Ego-VQA-1 & MCB & No Convergence & No Convergence \\
      Ego-VQA-2 & MCB & No Convergence & No Convergence \\\hline
      MSVD-QA & MCB & 30.85\% (-0.09) & 33.78\% (+0.36) \\\hline
      TGIF-Action & MCB & 73.56\% (+2.87) & 73.00\% (-0.87)\\
      TGIF-Count & MCB & 3.95$\dagger$ (0) & 3.98$\dagger$ (+0.06)\\
      TGIF-Trans & MCB & 79.30\% (+2.97) & 77.10\% (-1.84)\\
      TGIF-FrameQA & MCB & 51.72\% (-0.76) & 52.21\% (+0.80)\\
      
    \end{tabular}
    \caption{HME-VideoQA Model. The default fusion technique is concatenation. Dagger refers to minimised L2 loss.}
    \label{tab:cc2blp_hme}
  \end{center}
\end{table}

\section{Discussion}
We discuss our insights and speculations on our somewhat mixed results. Though we do not propose a cohesive theory for them, we offer speculation surveyed from surrounding multimodal literature.
\subsection{TVQA Experiments}
\noindent \textbf{Absolutely No BLP Improvements on TVQA: } On the HME concat-to-BLP substitution model (Table \ref{tab:cc2blp_hme}), MCB barely changes model performance at all. We find that none of our TVQA concat-to-BLP substitutions (Table \ref{tab:cc2blp_tvqa}) yield any improvements at all, with almost all of them performing worse overall (~0.3-5\%) than even the questionless concatenation model. Curiously, MCB scores the highest of all BLP techniques. The dual-stream model performs worse still, dropping accuracy by between ~5-10\% vs the baseline (Table \ref{tab:2stream}). Similarly, be find that MCB performs best despite it being the most dated BLP technique we trial that is known to require larger latent spaces to work on VQA.
\\\noindent \textbf{BERT Impacted the Most?: } For the TVQA BLP-substitution models, we find the GloVe, BERT and `no-subtitle' variations all degrade by roughly similar margins, with BERT models degrading more most often. This slight discrepancy is unsurprising as the most stable BERT baseline model is the best, and thus may degrade more on the inferior BLP variations. However, BERT's relative degradation is much more pronounced on the dual-stream models, performing 3\% worse than GloVe. We speculate that here, the significant and consistent drop is potentially caused by BERT's more contextual nature is no longer helping, but actively obscuring more pronounced semantic meaning learned from subtitles and questions.
\\\noindent \textbf{Blame Smaller Latent Spaces?: } Naturally, bilinear representations of time series data across multiple frames or subtitles are highly VRAM intensive. Thus we can only explore relatively small hidden dimensions (i.e. 1600). However, we cannot simply conclude our poor results are due to our relatively small latent spaces because: \textbf{I)} MCB is our best performing BLP technique. However, MCB has been outperformed by MFH on previous VQA models \textit{and} it has been shown to require much larger latent spaces to work effectively in the first place \cite{DBLP:journals/corr/FukuiPYRDR16} (~16000). \textbf{II)} Our vector representations of text and images are also much smaller (300-d) compared to the larger representation dimensions conventional in previous benchmarks (e.g. 2048 in \cite{DBLP:journals/corr/FukuiPYRDR16}). We note that 16000/2048 $\approx$ 1600/300, and so our latent-to-input size ratio is not radically different to previous works. 
\\\noindent \textbf{Unimodal Biases in TVQA and Joint Representation: } Another explanation may come from works exploring textual biases inherent in TVQA to textual modalities \cite{mbintvqa}. BLP has been categorised as a `joint representation'. \cite{Baltruaitis2019MultimodalML} consider representation as summarising multimodal data ``in a way that exploits the complementarity and redundancy of multiple modalities''. Joint representations combine unimodal signals into the same representation space. However, they struggle to handle missing data \cite{Baltruaitis2019MultimodalML} as they tend to preserve shared semantics while ignoring modality-specific information \cite{8715409}. The existence of unimodal text bias in TVQA implies BLP may perform poorly on the TVQA as a joint representation of it's features because: \textbf{I)} information from either modality is consistently missing, \textbf{II)} prioritising `shared semantics' over `modality-specific' information harms performance on TVQA. Though concatenation could also be classified as a joint representation, we argue that this observation still has merit. Theoretically, a concatenation layer can still model modality specific information, but a bilinear representation learns bilinear representations which would make modality specific information more challenging to learn. This may explain why our simpler BLP substitutions perform better than our more drastic `joint' dual-stream model.
\\\noindent \textbf{What About DCCA?: } Table \ref{tab:deepcca} shows our results on the DCCA augmented TVQA models. We see a slight but noticable performance degradation with this relatively minor alternation to the stream processor. As previously mentioned, DCCA is in some respects an opposite approach to multimodal fusion than BLP, i.e. a `coordinated representation'. The idea of a coordinated representations is to learn simultaneously learn a separate representation for each modality , but with respect to the other. In this way, it is thought that multimodal interactions can be learned while still preserving modality-specific information that a joint representation may otherwise overlook \cite{8715409, Peng2018ModalitySpecificCS}. DCCA specifically maximises cross-modal correlation. Without further insight from surrounding literature, it is difficult to conclude what TVQA's drop in performance using both joint \textit{and} coordinated representations could mean. We will revisit this when we discuss the role of attention in multimodal fusion.
\\\noindent \textbf{Does Content Matching Ruin Multimodal Integrity?: } The context matching technique used in the TVQA model is the birdirectional attention flow (BiDAF) module introduced in \cite{seo2016bidirectional}. It is used in machine comprehension between a textual context-query pair to generate query-aware context representations. BiDAF uses a `memoryless' attention mechanism where information from each time step does not directly affect the next, which is thought to prevent early summarisation. BiDAF considers different input features at different levels of granularity. The TVQA model uses bidirectional attention flow to create  context aware (visual/subtitle) question and answer representations, BiDAF can be seen as a co-ordinated representation in some regards, but it does project questions and answers representations into a new space. We use this technique to prepare our visual and question/answer features because it temporally aligns both features, giving them the same dimensional shape, conveniently allowing us to apply BLP at each time step. Since the representations generated are much more similar than the original raw features and there is some degree of information exchange, it may affect BLP's representational capacity. Though it is worth considering these potential shortcomings, we cannot immediately assume that BiDAF would cause serious issues as earlier bilinear technique were successfully used between representations in the same modality \cite{Tenenbaum:2000:SSC:1121517.1121518, DBLP:journals/corr/GaoBZD15}. This implies that multimodal interactions can still be learned between the more similar context-matched representations, provided the information is still present. Since BiDAF does allow visual information to be used in the TVQA baselline model, it is reasonable to assume that some of the visual information is in fact intact and exploitable for BLP. However, it is still currently unclear if context matching is fundamentally disrupting BLP and contributing to the poor results we find. We note that in BiDAF, `memoryless' attention is implemented to avoid propagating errors through time. We argue that though this may be true and help in some circumstances, conversely, this will not allow some useful interactions to build up over time steps.

\subsection{The Other Datasets on HME}
\noindent \textbf{BLP Has No Effect?: }
Our experiments on the Ego-VQA, TGIF-QA and MSVD-QA datasets are on concat-to-BLP substitution HME models. Frankly our results are inconclusive. There is virtually no variation in performance between the BLP and concatenation implementations. Interestingly, Ego-VQA consistently does not converge with this simple substitution. We cannot comment for certain on why this is the case. There seems to be no intuitive reason why it's $1^{st}$ person content would cause this. Rather, we believe this is symptomatic of overfitting in training, as Ego-VQA is very small \textit{and} pretrained on a different dataset. This would imply that BLP layers are more sensitive to related training difficulties.
\\\textbf{Does Better Attention Explain the Difference?: }
Attention mechanisms have been shown to improve the quality of text and visual interactions. \cite{yu2017multi} argue that methods without attention are `coarse joint-embedding models' which use global features that contain noisy information unhelpful in answering fine-grained questions commonly seen in VQA and video-QA. This is a strong motivation for implementing attention mechanisms alongside BLP, so that the theoretically greater representational capacity of BLP is not squandered on less useful noisy information. The TVQA model uses the previously discussed BiDAF mechanism to focus information from both modalities. However, the HME model integrates a more complex memory-based multi-hop attention mechanism. This difference may potentially highlight why the TVQA model suffers more substantially integrating BLP than the HME one. Further experimentation to explore this point would be very interesting.

\subsection{Neurological Parallels}
In this section we discuss how bilinear models in deep learning and multimodal fusion in general are related to 2 key areas of neurological research, i.e. the `two-stream' theory of vision \cite{goodale1992separate, Milner2017HowDT} and `dual coding' theory \cite{paivio2013imagery, Paivio2014IntelligenceDC}.
\\\noindent \textbf{Two-Stream Vision: } Introduced in \cite{goodale1992separate}, the current consensus on primate visual processing is that it is divided into two networks or streams: The `ventral' stream which mediates transforming the contents of visual information into `mental furniture' that guides memory, conscious perception and recognition, and the `dorsal' stream which mediates the visual guidance of action. There is a wealth of evidence showing that these two subsystems are not mutually insulated from eachother, but rather interconnect and contribute to one another at different stages of processing \cite{Milner2017HowDT, jeannerod2005visual}. In particular, \cite{jeannerod2005visual} argue that valid comparisons between visual representation must consider the direction of fit, direction of causation and the level of conceptual content. They demonstrate that visual subsystems and behaviours inherently rely on aspects of both streams. Recently, \cite{Milner2017HowDT} consider 3 potential ways these cross-stream interactions could occur: \textbf{I)} Computations along the 2 pathways are independent and combine at a `shared terminal' (the independent processing account), \textbf{II)} Processing along the separate pathways is modulated by feedback loops that transfer information from `downstream' brain regions, including information from the complementary stream (the feedback account), \textbf{III)} Information is transferred between the 2 streams at multiple stages and location along their pathways (the continuous cross-talk account). Though \cite{Milner2017HowDT} focus mostly on the `continuous cross-talk' idea, they believe that a unifying theory would include aspects from each of these scenarios. The vision-only deep bilinear models proposed in \cite{Tenenbaum:2000:SSC:1121517.1121518, Lin2015BilinearCF} are strikingly reminiscent to the $1^{st}$ `shared-terminal' scenario. The bilinear framework proposed in \cite{Tenenbaum:2000:SSC:1121517.1121518} focuses on splitting up `style' and `content', and is designed to be applied to any two-factor task. \cite{Lin2015BilinearCF} note but do not explore the similarities between their propose network and the two-stream model of vision. Their bilinear CNN model aims to processes two subnetworks separately, `what' (ventral) and `where' (dorsal) streams, and later combine in a bilinear `terminal'. BLP methods developed from these baselines would later focus on multimodal tasks between language and vision.
\\\noindent \textbf{Dual Coding Theory: } Dual coding theory (DCT) \cite{paivio2013imagery} broadly considers the interactions between the verbal and non-verbal systems in the brain (recently surveyed here \cite{Paivio2014IntelligenceDC}). DCT considers verbal and non-verbal interactions by way of `logogens' and `imagens' respectively, i.e. units of verbal and non-verbal recognition. Imagens may be multimodal, i.e. haptic, visual, smell, taste, motory etc. We should appreciate the distinction between medium and modality: image is both medium and modality and videos are an image based modality. Similarly, text is the medium through which the natural language modality is expressed. We can see parallels in multimodal deep learning and dual coding theory, with textual features as logogens and visual (and sometimes audio) features as visual (or auditory) imagens. There are many insights from DCT that could guide and drive multimodal deep learning: \textbf{I)} Logogens and imagens are discrete units of recognition and are often related to tangible concepts (e.g. `pictogens' \cite{Morton1979FacilitationIW}). This may imply that multimodal models should additionally focus on deriving more tangible features i.e. discrete convolution maps previously used in vision-only bilinear models \cite{Lin2015BilinearCF} as opposed to ImageNet-style feature vector more commonly used in recent BLP models and attention modules could be used to better visualise these learned relations. \textbf{II)} Multimodal cognitive behaviours in people can be improved by providing cues. For example, referential processing (naming an object or identifying an object from a word) has been found to additively affect free recall (recite a list of items), with the memory contribution of non-verbal codes (pictures) being twice that of verbal codes \cite{Paivio1981DualCA}. \cite{Begg1972RecallOM} find that free recall of `concrete phrases' (can be visualised) or their constituent words is roughly twice that of `abstract' phrases. However, this difference increased six-fold for concrete phrases when cued with one of the phrase words, yet using cues for abstract phrases did not help at all. This was named the `conceptual peg' effect in DCT, and is interpreted as memory images being re-activated by `a high imagery retrieval cue'. This may imply that future networks could improve in quality by focusing on learning referential relations between `concrete' words and images and treat `abstract' words and concepts differently. \textbf{III)} \cite{Bezemer2008WritingIM} explore the differences in student's understanding when text information is presented alongside other modalities. They argue that when meaning is moved from one medium to another semiotic relations are redefined. This paradigm could be emulated to control how networks learn concepts in relation to certain modal information. \textbf{IV)} Imagens (and potentially logogens) may be a function of many modalities, i.e. one may recognise something as a function of haptic and auditory experiences alongside visual ones. We believe this implies that non-verbal modalities (vision/sound etc..) should be in some way grouped or aggregated, and that while DCT remains widely accepted, multimodal research should consider `verbal vs non-verbal' interactions as a whole instead of focusing too intently on `case-by-case' interactions, i.e. text-vs-image and text-vs-sound. Recently proposed computational models of DCT have had many drawbacks \cite{Paivio2014IntelligenceDC}, we believe that neural networks are a natural fit for modelling neural correlates explored in DCT and should be considered as a future modelling option.

\subsection{Our Video-QA Fusion Recommendations}
We have experimented with BLP in 2 video-QA models and across 4 datasets. Though it is very difficult to draw strong conclusions from our experiments, we offer our recommendations for fusion informed by related multimodal surveys and neurological literature alongside our own experimental process: \textbf{I)} BLP as a fusion mechanism in video-QA can be exceedingly expensive due to added temporal relations. When using BLP for video-QA, we advise avoiding more computationally expensive variations across time, as memory limitations may force the hidden-size used sub-optimally low. Alternatively, summarising across time steps into condensed representations may allow more expensive and better BLP layers to be used. \textbf{II)} Our experiments, though limited, imply that there is little to gain in substituting concatenation steps between modalities for BLP in video-QA models. We recommend carefully choosing where the increased representational power is needed and not blindly replacing concatenation steps. \textbf{III)} Attention mechanisms are pivotal in VQA for reducing noise and focusing on specific fine-grained details \cite{yu2017multi}. We believe the sheer increase in feature information moving from still-image to video further increases the importance of attention in video-QA. The HME model was not as degraded by BLP as the TVQA one, as discussed this may be due to more sophisticated attention mechanisms better directing the BLP layer. \textbf{IV)} BLP in HME consistently failed to converge on Ego-VQA, a smaller but specialised dataset. We advise caution in using BLP methods that are sensitive to training hyper-parameters (MLB) in scenarios such as Ego-VQA that will already be more difficult to train due to the required pretraining and smaller sizes. \textbf{V)} \cite{Long2018MultimodalKA} consider multiple different fusion methods for video classification, i.e. LSTM, probability, `feature' and attention. `Feature' fusion is the direct connection of each modality within each local time interval, which is essentially what context matching does in the TVQA model. \cite{Long2018MultimodalKA} finds temporal feature based fusion sub-par, and speculates that the burden of learning multimodal \textit{and} temporal interactions is too heavy. It is instead recommended that for video classification, attention based fusion is best. Our poor TVQA results similarly may imply that modelling temporal and multimodal relations via BLP is too difficult for video-QA. We therefore recommend focusing primarily on attention based fusion when considering temporal relations in TVQA, inline with the findings in \cite{Long2018MultimodalKA}. \textbf{VI)} Perhaps most importantly, \textbf{we recommend reading into related neurological theories when working on multimodal deep learning.} DCT and two-stream models offer a wealth of insights into the nature of human cognition that will inevitably be very helpful for any future truly multimodal neural network based AI.

\subsection{Proposed Areas of Research: }
We outline some potential future research directions our efforts have revealed. \textbf{I)} More experimental results. Our experiments are too limited to draw any meaningful conclusions, extra experiments with complementing or contradictory results will allow the field greater insight into the potential role of BLP in video-QA. \textbf{II)} Discrete feature maps. We believe that bilinear fusion techniques have strayed from more concrete convolution maps into more generalised ImageNet-style feature vectors. From the logogen-imagen based insights discussed about DCT, we believe experiments contrasting generalised feature vectors with more tangible ones would offer a great deal of insight. \textbf{III)} Bilinear representations are not the most complex ways to learn interactions between modalities, as explored in \cite{Yu2018BeyondBG}, we believe that higher-order interactions between features will facilitate more realistic models of the world. We would like to note the non-linear extension of BLP, in particular bi-nonlinear interactions (i.e. non-linear in the same manner with respect to both its input) would increase the representational capacity of bilinear models further.

\section{Conclusion}
In light of BLP's success in VQA, we have experimentally explored their use in video-QA on 2 model and 4 datasets. We find that switching from vector concatenation to BLP through simple substitution on the HME and TVQA models do not improve and in fact actively harm performance on video-QA. We find that a more substantial `dual-stream' restructuring of the TVQA model to accommodate BLP significantly reduces performance on TVQA. Our experiments results imply that naively using BLP techniques can be very detrimental in video-QA. We caution against automatically integrating bilinear pooling in video-QA models and expecting similar results to still-image-QA. We offer several interpretations and insights of our negative results using surrounding multimodal and neurological literature and find our results inline with trends in VQA and video-classification. To the best of our knowledge, we are the first to extensively outline how important neurological theories i.e. dual coding theory and the two-stream model of vision relate to modern deep learning practices. We offer a few experimentally and theoretically guided tips to consider for multimodal fusion in video-QA, most notably that attention mechanisms should be prioritised over BLP or other direct feature fusion techniques. We would like to emphasise the importance of related neurological theories in deep learning and encourage researchers to explore dual coding theory and the two-stream model of vision.

\bibliographystyle{IEEEtran}
\bibliography{mybibs}
\end{document}